\documentclass[sigconf]{acmart}

\renewcommand\footnotetextcopyrightpermission[1]{}
\usepackage{graphicx}
\usepackage{booktabs}
\usepackage{multirow}
\usepackage{array}
\usepackage{float}
\usepackage{amsmath}

\usepackage{amssymb}
\usepackage{colortbl}
\usepackage{tcolorbox}
\usepackage{adjustbox}
\usepackage{enumitem}
\usepackage{fdsymbol}
\usepackage{algorithm}
\usepackage{algorithmic}
\usepackage{hyperref}

\AtBeginDocument{%
  \providecommand\BibTeX{{%
    \normalfont B\kern-0.5em{\scshape i\kern-0.25em b}\kern-0.8em\TeX}}}

\begin{document}

\title{Thinking Before Matching: A Reinforcement Reasoning Paradigm Towards General Person Re-Identification}


\author{Quan Zhang}
\authornote{Equal contribution}
\affiliation{%
  \institution{Sun Yat-sen University}
  }
\email{zhangq689@mail.sysu.edu.cn}

\author{Jingze Wu}
\authornotemark[1]
\affiliation{
  \institution{Sun Yat-sen University}
  }
\email{wujz3@mail2.sysu.edu.cn}

\author{Jialong Wang}
\affiliation{
  \institution{Alibaba Cloud Computing}
  }
\email{quming.wjl@alibaba-inc.com}

\author{Xiaohua Xie}
\affiliation{
  \institution{Sun Yat-sen University}
  }
\email{xiexiaoh6@mail.sysu.edu.cn}

\author{Jianhuang Lai}
\authornote{Corresponding author}
\affiliation{
  \institution{Sun Yat-sen University}
  }
\email{stsljh@mail.sysu.edu.cn}

\author{Hongbo Chen}
\affiliation{
  \institution{Sun Yat-sen University}
  }
\email{chenhongbo@mail.sysu.edu.cn}


\begin{abstract}
Learning identity-discriminative representations with multi-scene generality has become a critical objective in person re-identification (ReID). However, mainstream perception-driven paradigms tend to identify fitting from massive annotated data rather than identity-causal cues understanding, which presents a fragile representation against multiple disruptions. 
In this work, ReID-R is proposed as a novel reasoning-driven paradigm that achieves explicit identity understanding and reasoning by incorporating chain-of-thought into the ReID pipeline. Specifically, ReID-R consists of a two-stage contribution: (i) Discriminative reasoning warm-up, where a model is trained in a CoT label-free manner to acquire identity-aware feature understanding; and (ii) Efficient reinforcement learning, which proposes a non-trivial sampling to construct scene-generalizable data. On this basis, ReID-R leverages high-quality reward signals to guide the model toward focusing on ID-related cues, achieving accurate reasoning and correct responses. Extensive experiments on multiple ReID benchmarks demonstrate that ReID-R achieves competitive identity discrimination as superior methods \textbf{using only 14.3K non-trivial data (20.9\% of the existing data scale)}. Furthermore, benefit from inherent reasoning, ReID-R can provide high-quality interpretation for results.
\end{abstract}


\maketitle

\begin{figure}[!t]
    \centering
\includegraphics[width=1.0\linewidth]{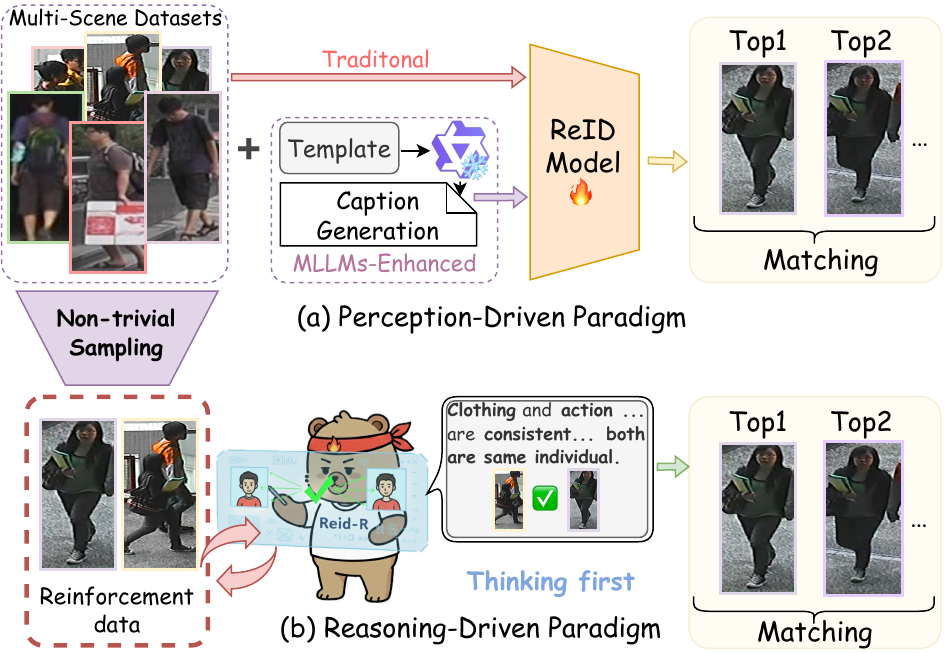}
    \caption{\textbf{Perception-Driven vs. Reasoning-Driven ReID Paradigms.} (a) Mainstream \textbf{Perception-Driven} methods (traditional or MLLMs-enhanced methods) are trained on \textbf{massive, multi-scene datasets}. They tend to learn fitting rather than understanding identity-causal cues, resulting in a fragile situation against disruptions. (b) Our proposed \textbf{Reasoning-Driven} paradigm (ReID-R) incorporates an explicit chain-of-thought process to understand these identity-causal cues, after being trained on our Non-trivial sampling dataset, achieving both robust matching and high-quality interpretation.}
    \label{fig:intro}
\end{figure}

\section{Introduction}

Person Re-Identification (Re-ID) aims to match the same individual across non-overlapping camera views and serves as a fundamental problem in intelligent surveillance and public-security applications~\cite{deep-reid-1, deep-reid-2,deep-reid-3,deep-reid-4,deep-reid-7,deep-reid-8}. As the field advances, the goal of Re-ID has increasingly shifted toward building a single model that can reliably discriminate identities and generalize across diverse scenes, where most existing approaches still belong to perception-driven modeling. However, both traditional deep learning paradigms~\cite{deep-reid-5,deep-reid-6,deep-reid-9,deep-reid-10,deep-reid-11} and methods leveraging Multimodal Large Language Model (MLLM)-enhanced inputs~\cite{llm-reid-1,llm-reid-2,llm-reid-3,llm-reid-4, bai2023qwen,wang2024qwen2,liu2024visual,yu2025eve} (Fig.~\ref{fig:intro}(a)) rely heavily on large-scale annotated datasets. By fitting the statistical distributions behind massive datasets, perception-driven models implicitly capture identity features. However, they remain inherently vulnerable to scene-specific distractors because they rely on implicit pattern matching rather than explicit deduction. This limitation motivates the need to move toward a next-level paradigm capable of accurately and efficiently understanding and reasoning about identity-related cues.

Reinforcement learning (RL) has recently demonstrated strong effectiveness in equipping models with chain-of-thought reasoning (CoT), rapidly attracting attention in multimodal fine-tuning reasoning~\citep{xiao2025fastslow,wu2025spatialmllm,wang2025timer,peng2025lmm,zhan2025visionr1,zhang2025r1,zhou2025r1}. However, directly transferring RL fine-tuning to Re-ID is infeasible, because the original foundation model lacks a basic concept of ``identity'', causing its reasoning to rely on unreliable visual cues. Specifically, without sufficient awareness of which features are critical for Re-ID, the model tends to exploit data biases to “cheat” the reward signal, ultimately resulting in poor reasoning optimization~\cite{guo2025deepseek,videor1,wang2025videorft}.

Based on the above analysis, ReID-R (i.e., ReID-Reasoning) is proposed as a novel reasoning-driven paradigm. As shown in Fig.~\ref{fig:intro}(b), ReID-R is a pure reinforcement learning framework that introduces CoT into ReID, enabling identity understanding and reasoning. By integrating knowledge from MLLMs, such a paradigm can identify stable identity features with significantly less training data (e.g., using only 20.9\% of the standard data scale), achieving robust performance even in challenging scenarios. Specifically, ReID-R consists of a two-stage contribution: (i) Discriminative Reasoning Warm-up, which introduces a novel ``Discriminative Captioning'' task, the policy model guided by a novel contrastive reward to autonomously discover salient and generalizable identity ``evidence'' in a CoT label-free manner, thus acquiring robust identity-aware feature understanding. (ii) Efficient Reinforcement Learning, which designs a non-trivial sampling method to solve the optimization challenge of reward variance. This filtering process isolates a potent, learnable subset from the data, ensuring a rich gradient signal for the RL optimizer. On this foundation, the model is trained to leverage the ``evidence'' from Stage one, focusing on ID-related cues to achieve accurate reasoning and correct responses.

Extensive experiments to validate the superiority of our framework. Our results across five challenging benchmarks show that ReID-R achieves highly competitive performance compared to SOTA black-box models, using only 14.3K nontrivial data points (representing merely 20.9\% of the existing data scale). Crucially, our qualitative analyses demonstrate the model's high interpretability as shown in Fig.~\ref{fig:intro}(b), demonstrating that its decisions are grounded in verifiable, logical justifications. We further show that the robustness of the features learned in our warm-up stage allows the model to focus on identity-persistent cues (e.g., ``posture'') rather than superficial distractors (e.g., ``clothing color''), proving the effectiveness of our two-stage design. In practical deployments, this capability to explicitly reason through challenging real-world samples provides verifiable decision support and significantly enhances overall system transparency.

Our contributions are as follows:
\begin{itemize}[leftmargin=*] 
\item We propose a novel reasoning-driven paradigm for person re-identification and design the \textbf{ReID-R }framework, which first designs a non-trivial sampling strategy to construct high-quality reinforcement data, and then proposes a two-stage reinforcement learning strategy to achieve identity understanding and CoT reasoning.

\item We design a Discriminative Reasoning Warm-up to achieve identity understanding in a CoT label-free manner, which formalizes identity understanding into a ``Discriminative Captioning'' task. Through designing a ReID-inspired reward, our model is guided to generate CoT that cover more critical identity components, which serve as robust supporting evidence for the subsequent reasoning stage.

\item Extensive experiments demonstrate that ReID-R, powered by a reasoning-driven paradigm, achieves competitive identity discrimination using only 14.3K non-trivial samples (20.9\% of the original data). Moreover, this paradigm enables ReID-R with high-quality interpretability for matching results.
\end{itemize}

\section{Related Works}

\subsection{Person Re-Identification}
\label{sec:reid_related_work}
Discriminative identity learning has long been a fundamental challenge in person re-identification~\cite{ye2021deep, zheng2016person, zheng2017person, yuan2020defense, hermans2017defense}.
As shown in Fig.~\ref{fig:intro}(a), the dominant research line follows a perception-driven paradigm, encompassing both traditional methods and recent MLLM-enhanced approaches. Traditional research primarily focuses on optimizing performance within a single scenario and has demonstrated strong results in domains such as standard Re-ID~\cite{zheng2017person, hermans2017defense} and clothes-changing Re-ID~\cite{huang2021clothing, gu2022clothes}. However, these methods often degrade when applied to cross-domain settings. The rise of  MLLM, trained on massive multi-scene datasets, provides a new path toward improved generalization through richer semantic understanding. Accordingly, several recent works integrate MLLMs to refine or augment input features~\cite{li2023clip, yang2024pedestrian, he2024instruct, lu2025llavareid, wang2025idea}. Despite their promising improvements, such MLLM-enhanced methods remain within the perception-driven paradigm: they still rely on fitting statistical patterns in large-scale data to implicitly capture identity features and generate matching lists, rather than performing explicit reasoning for identity understanding.

In contrast, our approach proposes a reasoning-driven modeling paradigm (Fig.~\ref{fig:intro}(b)). By integrating knowledge from MLLMs, we introduce a pure reinforcement learning framework for ReID that leverages CoT to explicitly identify stable identity features. This enables the model to understand and infer identities robustly using only 14.3K non-trivial samples (20.9\% of the original data).

\subsection{Enhancing MLLMs Reasoning}
\label{sec:related_work_cot}

Reinforcement learning (RL)-based fine-tuning has become the dominant paradigm for acquiring complex reasoning skills~\cite{xiao2025fastslow, wu2025spatialmllm, wang2025timer, peng2025lmm, zhan2025visionr1, zhang2025r1, zhou2025r1}, especially after the success of DeepSeek-R1~\cite{guo2025deepseek}. Unfortunately, directly applying RL to ReID is infeasible, because RL methods need a ``cold start'' to understand identity in ReID. Therefore, existing RL methods in other domains try to collect task-specific annotations and pre-performing Supervised Fine-Tuning (SFT), such as Vision-R1~\cite{huang2025vision}, Video-R1~\cite{videor1}, LMM-R1~\cite{peng2025lmm}. However, this strategy is impractical for ReID: collecting identity-related reasoning annotations at scale is unaffordable, and SFT on noisy or limited annotations can lead to overfitting and poor generalization~\cite{guo2025deepseek, wang2025videorft}.

These limitations motivate our new paradigm. Instead of relying on general-purpose SFT reasoning data, ReID-R proposes a discriminative reasoning warm-up for ID understanding in a CoT label-free manner. Furthermore, ReID-R performs efficient reinforcement learning to achieve ID reasoning. Compared with traditional SFT, ReID-R enables effective reasoning under minimal supervision.

\section{Methodology}

\subsection{Preliminary}
\label{subsec:Preliminary}
In traditional person ReID, the goal is typically to learn a discriminative embedding model. This model encodes a query image $I_q$ and a large set of gallery images $\{I_g^1, I_g^2, \dots, I_g^N\}$ into feature vectors. A ranked list is then produced by computing the similarity between the query vector and all gallery vectors. However, our goal is to leverage MLLMs for interpretable ReID. Due to the limited context space of MLLMs, processing the entire gallery set $\{I_g^i\}$ simultaneously with the query is often infeasible. Therefore, we reformulate the task as a pairwise decision problem, focusing exclusively on the top-$k$ hard candidates. This allows us to leverage MLLM reasoning to boost accuracy on complex samples and generate explicit reasoning traces for transparent decision support.

To begin with, we formulate the person ReID task as a multi-round decision process. Given a query image $I_q$ and a gallery candidate $I_g$, the goal is to optimize a policy model $\pi_{\theta}$ that can generate a reasoning trace $o$ and lead to the selection of the correct binary decision $a^{*} \in \{0,1\}$. The model $\pi_{\theta}$ autoregressively generates a sequence $\mathcal{O}=[t_1,t_2,\dots,t_{L}]$ which includes both the reasoning steps (e.g., comparing attributes) and the final decision. The objective of the RL fine-tuning is to find the optimal parameters $\theta$ that maximize the expected reward $R$ for the generated traces:
\begin{equation} \mathcal{J}(\theta) = \mathbb{E}_{(I_q, I_g)\sim\mathcal{D}, \mathcal{O}\sim\pi_{\theta}(\cdot|I_q, I_g)} \left[ R(\mathcal{O}) \right]. \label{eq:rl_objective}
\end{equation}

For optimizing the policy model $\pi_{\theta}$, our work builds upon the GRPO framework~\cite{shao2024deepseekmath}: for a given query-gallery pair $(I_q, I_g)$, the policy model $\pi_{\theta}$ generates a set of $G$ candidate reasoning traces $\{o_1,\dots,o_G\}$. A reward model then assigns each trace $o_i$ a score $R_i$. For example, rule-based rewards usually include a format reward $\mathcal{R}_{format}$ and an accuracy reward $\mathcal{R}_{acc}$:
\begin{equation}
\label{eq:baseline_reward}
\mathcal{R}_{acc} = \mathbb{I}(a_i = a^*),
\end{equation}
where $a_i$ represents the final decision within trace $o_i$ and $\mathbb{I}(\cdot)$ is the indicator function evaluating against the ground-truth $a^*$. From this set of $G$ scores, a standardized advantage $\hat{A_i}$ is computed for each trace. This is achieved by normalizing the individual scores against the group's mean and standard deviation:
\begin{equation}
\label{eq:ro}
    \hat{A_i}=
    \frac{R_i-\mathrm{mean}(\{R_i\}_{g=1}^G)}{\mathrm{std}(\{R_i\}_{g=1}^G)} .
\end{equation}
The policy $\pi_\theta$ is subsequently trained by maximizing the following objective function:
\begin{equation}
\begin{aligned} \mathcal{J}_\text{GRPO}(\theta) =& \mathbb{E}_{(I_q, I_g)\sim \mathcal{D}, \{o_g\}_{g=1}^G\sim \pi_{\theta_\text{old}}(\cdot\mid I_q, I_g)} \\& \Bigg[ \frac{1}{G}\sum_{g=1}^{G} \frac{1}{|o_g|}\sum_{t=1}^{|o_g|} \Bigg( \min \Big( r_{g,t}(\theta) \hat{A}_{g,t}, \\& \ \text{clip} \Big( r_{g,t}(\theta), 1 - \varepsilon, 1 + \varepsilon \Big) \hat{A}_{g,t} \Big) \\&- \beta D_{\text{KL}}(\pi_{\theta} || \pi_{\text{ref}}) \Bigg) \Bigg], 
\label{eq:grpoloss}
\end{aligned}
\end{equation}
where $r_{i,t}(\theta)$ denotes the importance sampling ratio between the new and old policies~\cite{shao2024deepseekmath}, and the $D_{\text{KL}}$ term acts as a penalty to constrain policy updates and ensure stability.

However, Person Re-ID is a challenging fine-grained retrieval task where identity is a highly abstract concept. We find that applying the GRPO framework (Eq. \ref{eq:grpoloss}) directly to a general-purpose MLLM yields suboptimal results. The core issue is that the base model $\pi_{\theta}$ itself lacks the foundational knowledge of Re-ID. It struggles to identify the subtle, discriminative features required for accurate reasoning, getting ``lost'' in the vast semantic space. It cannot effectively utilize the simple accuracy reward (Eq. \ref{eq:baseline_reward}) because it doesn't know what features to focus on to achieve that accuracy.

To solve this, we propose ReID-R, a reinforcement-based reasoning framework designed as a two-stage training strategy. As shown in Fig.~\ref{fig:reidr1_overview}, our strategy is first to teach the model what effective identity features are, and then teach it how to reason.

\begin{figure*}[t]
    \centering
    \includegraphics[width=1.0\linewidth]{reidr1_framework.pdf}
    \caption{The pipeline of ReID-R. (a) Discriminative Reasoning Warm-up for ID Understanding: we first train a model to generate identity-aware and discriminative captions. (b) Step 2: Efficient Reinforcement Learning for ID Reasoning: we reformulate the ReID task as a multi-round decision-making process. The model generates explicit reasoning chains (e.g., comparing attributes such as age and clothing) to make a final decision.}
    \label{fig:reidr1_overview}
\end{figure*}

\subsection{Reid-R Paradigm}
\label{subsec:Reid-R1}
\subsubsection{Overview}
Our framework consists of two key stages: (i) {Discriminative Reasoning Warm-up for ID Understanding} and (ii) {Efficient Reinforcement Learning for ID Reasoning}. First, Discriminative reasoning warm-up trains a specialized model to generate a diverse set of identity-aware captions. This provides the essential, high-quality descriptive ``evidence'' required for the subsequent complex reasoning task. Subsequently, Efficient reinforcement learning applies the GRPO objective to this ``warm-up'' model, training it to generate explicit step-by-step comparisons and make a verifiable final prediction. The synergy of high-quality preparation and robust reasoning optimization allows our model to achieve both accuracy and interpretability.

\subsubsection{Stage 1: Discriminative Reasoning Warm-up for ID Understanding}

\begin{figure}[t]
    \centering
    \includegraphics[width=\linewidth]{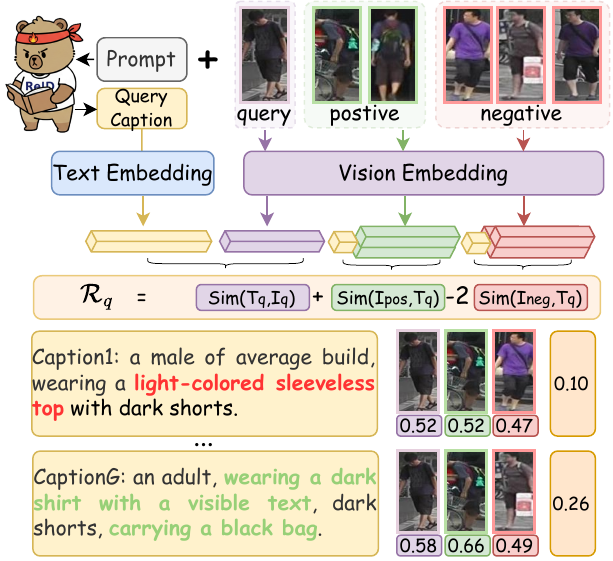}
    \caption{Illustration of the query anchored contrastive reward. The reward $\mathcal{R}_q$ is calculated by measuring the alignment between the generated query caption $T_q$ and the visual embeddings of the query $I_q$, the positive sample $I_{\text{pos}}$, and the negative sample $I_{\text{neg}}$. A caption that better discriminates the query person from non-matches receives a higher reward, guiding the model to learn identity-aware features.}
    \label{fig:vis}
\end{figure}

For any reasoning task, the quality of the final judgment depends on the quality of the initial premises, or evidence. For Re-ID, this ``evidence'' is the set of discriminative identity features extracted from the person's image. However, as shown in Fig.~\ref{fig:ablation_viz}, we find that traditional MLLMs do not inherently understand which identity features are stable. Directly applying RL to a base model causes it to mistakenly rely on changeable identity attributes during reasoning, ultimately leading to reasoning failure. To enable the MLLMs to extract these crucial features, a common approach is to fine-tune them on an image captioning task. But standard captioning methods typically rely on a reward signal computed by matching the generated text against a ground-truth caption. And this method presents two major limitations. First, scaling fine-grained, identity-aware annotations incurs prohibitive costs. Second, standard captions typically describe images in isolation, depriving the model of the contrastive signals necessary to disambiguate visually similar individuals. 

To overcome this limitation and teach the model to extract truly identity features, we design a novel task: \textbf{Discriminative Captioning}. As shown in Fig.~\ref{fig:vis}, this task does not require any captions, which guides the model to autonomously discover identity features using a novel contrastive reward. Specifically, our task is formulated as a contrastive multiple-choice problem. The model is presented with a query image alongside a set of gallery images ($K \ge 2$), explicitly including both positive (matching identity) and negative (non-matching identity) samples. Instead of a direct prediction, the objective is to first generate intermediate textual observations that describe each image, and subsequently synthesize them into a final discriminative conclusion. An effective observation should semantically capture the commonalities between the query and positive samples while explicitly isolating them from the negative samples.

Then, the quality of these generated intermediate observations is evaluated by a \textbf{Multi-Perspective Contrastive Reward}. We use a pre-trained SigLIP model~\cite{zhai2023sigmoid} to encode the generated captions into textual vectors ($T_q, T_{\text{pos}}, T_{\text{neg}}$) and their corresponding images into visual vectors ($I_q, I_{\text{pos}}, I_{\text{neg}}$). The contrastive reward $\mathcal{R}_{\text{c}}$ is the average of three components:
\begin{equation}
\mathcal{R}_{\text{c}} = \frac{1}{3} \left( \mathcal{R}_{q} + \mathcal{R}_{\text{pos}} + \mathcal{R}_{\text{neg}} \right).
\label{eq:obs_reward}
\end{equation}
Specifically, each component evaluates the alignment anchored on its respective text perspective. We define the contrastive objectives for the query ($\mathcal{R}_{q}$), positive ($\mathcal{R}_{\text{pos}}$), and negative ($\mathcal{R}_{\text{neg}}$) anchors as follows:
\begin{equation}
\begin{aligned}
\mathcal{R}_{q} &= \operatorname{sim}(T_q, I_q) + \operatorname{sim}(T_q, I_{\text{pos}}) - 2\operatorname{sim}(T_q, I_{\text{neg}}), \\
\mathcal{R}_{\text{pos}} &= \operatorname{sim}(T_{\text{pos}}, I_{\text{pos}}) + \operatorname{sim}(T_{\text{pos}}, I_q) - 2\operatorname{sim}(T_{\text{pos}}, I_{\text{neg}}), \\
\mathcal{R}_{\text{neg}} &= 2\operatorname{sim}(T_{\text{neg}}, I_{\text{neg}}) - \operatorname{sim}(T_{\text{neg}}, I_{\text{pos}}) - \operatorname{sim}(T_{\text{neg}}, I_q),
\end{aligned}
\label{eq:tripartite_reward}
\end{equation}
where $\operatorname{sim}(\cdot, \cdot)$ denotes the cosine similarity. This formulation symmetrically penalizes incorrect cross-modal associations and encourages the model to isolate features that separate identities.

Moreover, to ensure the model genuinely extracts and applies the extracted identity information, we introduce a gating mechanism: the model obtains the contrastive reward only when it correctly judges which gallery image matches the query identity. The final reward $\mathcal{R}$ is formulated as:
\begin{equation}
R =
\begin{cases}
R_\text{c}+R_\text{acc}+R_\text{format}, & \text{if }  \mathcal{R}_{acc}=1, \\
R_\text{format}, & \text{if } \mathcal{R}_{acc}=0.
\end{cases}
\label{eq:final_reward}
\end{equation}

\subsubsection{Stage 2: Efficient Reasoning Optimization for ID Reasoning}
While Stage One successfully primes the model to extract high-quality identity features as robust ``evidence,'' and deduce the matching identity under the strong prior that a positive sample exists within the gallery. But in practical applications, the input gallery typically lacks such a prior. To adapt the model to real world applications, this stage transitions to an independent pairwise verification task. Specifically, the model is required to directly compare the query image ($I_q$) and a single candidate gallery image ($I_c$) to determine whether they share the same identity. To ensure the generated reasoning remains factually grounded to their respective visual inputs during this comparison, the contrastive reward $\mathcal{R}_{\text{c}}$ is designed to align each text description with its corresponding image:
\begin{equation}
\mathcal{R}_{\text{c}} = \frac{1}{2} \big( \operatorname{sim}(T_q, I_q) + \operatorname{sim}(T_c, I_c) \big),
\label{eq:stage2_contrastive_reward}
\end{equation}
where $T_q$ and $T_c$ denote the generated textual observations for the query and candidate images, respectively. Finally, the final reward $\mathcal{R}$ is formulated as:
\begin{equation}
\mathcal{R} =
\begin{cases}
\mathcal{R}_{\text{c}} + \mathcal{R}_{\text{acc}} + \mathcal{R}_{\text{format}}, & \text{if } \mathcal{R}_{\text{acc}} = 1, 
\\
\mathcal{R}_{\text{format}}, & \text{if } \mathcal{R}_{\text{acc}} = 0.
\end{cases}
\label{eq:stage2_final_reward}
\end{equation}

During training, to ensure the generalization of our model, we merge training sets from several public benchmarks (e.g., Market1501~\cite{zheng2015person}, MSMT17~\cite{wei2018person}), resulting in a diverse dataset of 68,445 images. In conventional ReID training, an input query requires comparison against all images in the gallery. Exhaustively pairing all these images creates an intractable search space of over 4.6 billion pairs. Performing online reinforcement learning inference across this massive space for MLLMs is computationally prohibitive. Furthermore, since the vast majority of these pairs are trivial negative samples, training on such an imbalanced distribution provides weak gradient signals and ultimately leads to reward collapse.

To overcome this scalability bottleneck, we abandon naive pair generation and propose the \textbf{Non-trivial Data Sampling}. The core intuition is that hard negative samples are the most harmful elements in retrieval tasks. In the feature space of conventional models, these hard negatives are deceptively close to easy positives, causing misranking and severely degrading Rank-1 accuracy. Fortunately, pretrained MLLMs possess general knowledge and can deduce identity at a semantic level. Therefore, we aim to explicitly leverage this capability to tackle the hard negative challenge.

Specifically, instead of utilizing the full gallery, we construct highly informative reasoning triplets to better simulate real world application demands while ensuring training efficiency. In practice, we utilize a standard ViT-based retrieval model, independently trained on each respective dataset, to perform initial retrieval. For each query $I_q$, we retrieve the Top K most similar images from the gallery, forming a restricted subset $\mathcal{G}_{K}(I_q)$. We randomly select queries where this Top K subset contains at least one negative sample. Let $\mathcal{P}_{K}(I_q)$ and $\mathcal{N}_{K}(I_q)$ denote the sets of positive and negative samples within this retrieved subset, defined as:
\begin{equation}
\begin{aligned}
    \mathcal{P}_{K}(I_q) &= \{ I \in \mathcal{G}_{K}(I_q) \mid y(I) = y(I_q) \}, \\
    \mathcal{N}_{K}(I_q) &= \{ I \in \mathcal{G}_{K}(I_q) \mid y(I) \neq y(I_q) \},
\end{aligned}
\end{equation}
where $y(\cdot)$ denotes the ground truth identity label. Given that a standard training gallery exceeds 1,000 images, we define any samples appearing within this narrow Top K ranking as inherently easy positives and hard negatives. By randomly sampling an easy positive $I_p \in \mathcal{P}_{K}(I_q)$ and a hard negative $I_n \in \mathcal{N}_{K}(I_q)$, we formulate a challenging triplet $\langle I_q, I_p, I_n \rangle$. This targeted strategy forms our Candidate Pool $\mathcal{D}_{\text{pool}}$:
\begin{equation}
\mathcal{D}_{\text{pool}} = \left\{ \langle I_q, I_p, I_n \rangle \mid I_p \in \mathcal{P}_{K}(I_q), I_n \in \mathcal{N}_{K}(I_q) \right\}.
\end{equation}
In this way, 4.6 billion potential pairs are reduced to 14,300 distinct and challenging pairs. Specifically, as observed in Tab.~\ref{OmniReID benchmark}, although the initial image counts vary significantly from 7,365 in CUHK03 to over 30,000 in MSMT17, the final sample contributions remain highly consistent at approximately 3,500 per dataset. By dynamically adjusting the sampling rate for each source, our NTS strategy prevents the training process from being dominated by any single large scale dataset. This equilibrium ensures that the model learns representative identity features from all scenarios equally, which effectively eliminates simple samples and forms a high quality foundation for our training.

\begin{table}[t]
    \caption{Overview of ReID-R training data preparation by the proposed NTS strategy from multiple datasets.}
    \label{OmniReID benchmark}
    \centering
    \begin{tabular}{lccc} 
    \toprule
    Dataset    & \#Image & \#NTS\%   & Samples                    \\ 
    \midrule
    MSMT17~\cite{wei2018person}     & 30,248  & 11.7 & 3,547       \\
    CUHK03~\cite{li2014deepreid}       & 7,365 &49.2 & 3.626                   \\
    Market1501~\cite{zheng2015person} & 12,936 &27.6 & 3,576                  \\
    PRCC~\cite{yang2019person}       & 17,896 &19.8 & 3,551                     \\
    \midrule
    Ours & - & 20.9 &  14,300  \\
    \bottomrule
    \end{tabular}
\end{table}

\begin{table}[t]
\centering
\setlength{\tabcolsep}{4pt}
\renewcommand\arraystretch{1.0}
\caption{Comparison of ReID-R with SOTA methods across three traditional person Re-ID datasets.}
\begin{tabular}{lcccccc}
\toprule
\multicolumn{1}{l}{\multirow{2}{*}{\textbf{Methods}}}
&\multicolumn{2}{c}{\textbf{Market1501}} &\multicolumn{2}{c}{\textbf{MSMT17}} &\multicolumn{2}{c}{\textbf{CUHK03}}  \\
\cline{2-7} 
 &mAP & Rank-1& mAP & Rank-1& mAP & Rank-1 \\
\toprule
TransReID~\cite{he2021transreid} &86.8 & 94.4 & 61.0 & 81.8 & - & -\\
SAN~\cite{jin2020semantics}&88.0 & 96.1 & - & - & 76.4 & 80.1 \\
HumanBench~\cite{tang2023humanbench}&89.5 & - & 69.1 & - & 77.7 & -\\
PASS~\cite{zhu2022pass}&\textbf{93.0} & \textbf{96.8} &\underline{ 71.8 }& \textbf{{88.2}} & - & -\\
IRM~\cite{he2024instruct}& {92.3} & {96.2} & \textbf{{71.9}} & 86.2 & \underline{83.3 }& \underline{{86.5}} \\
\hline
ReID-R & \underline{92.5} & \underline{96.5} & {69.3} & \underline{{86.8}} & {\textbf{85.6}}& {\textbf{87.7}} \\
\bottomrule
\end{tabular}%
\label{tab:tra-ReID}%
\end{table}%

\begin{table}[t]
  \centering
  \renewcommand\arraystretch{1.0}
  \caption{Comparison of ReID-R with SOTA methods across two cloth-changing Re-ID datasets.}
    \begin{tabular}{lcccc}
    \toprule
    \multicolumn{1}{l}{\multirow{2}{*}{\textbf{Methods}}} &\multicolumn{2}{c}{\textbf{PRCC}} &\multicolumn{2}{c}{\textbf{VC-Clothes}} \\
    \cline{2-5} & mAP & Rank-1 & mAP & Rank-1\\
    \hline
    RGA-SC~\citep{zhang2020relation} & - & 42.3 & 67.4 & 71.1\\
    PCB~\citep{sun2018beyond} & 38.7 & 41.8 & 62.2 & 62.0\\
    IANet~\citep{hou2019interaction} & 45.9 & 46.3 & - & -\\
    TransReID~\citep{he2021transreid} & - & 44.2 & 71.8 & 72.0\\
    IRM~\cite{he2024instruct} & 46.0 & \textbf{48.1} & \textbf{80.1} & \underline{90.1} \\
    \hline
    ReID-R & \textbf{46.2} & {\underline{46.4}} & \underline{78.8} & \textbf{90.2} \\
    \toprule
    \end{tabular}%
  \label{tab:CC-ReID}%
\end{table}%

\section{Experiment}

\begin{figure*}[!t]
    \centering
    \includegraphics[width=\linewidth]{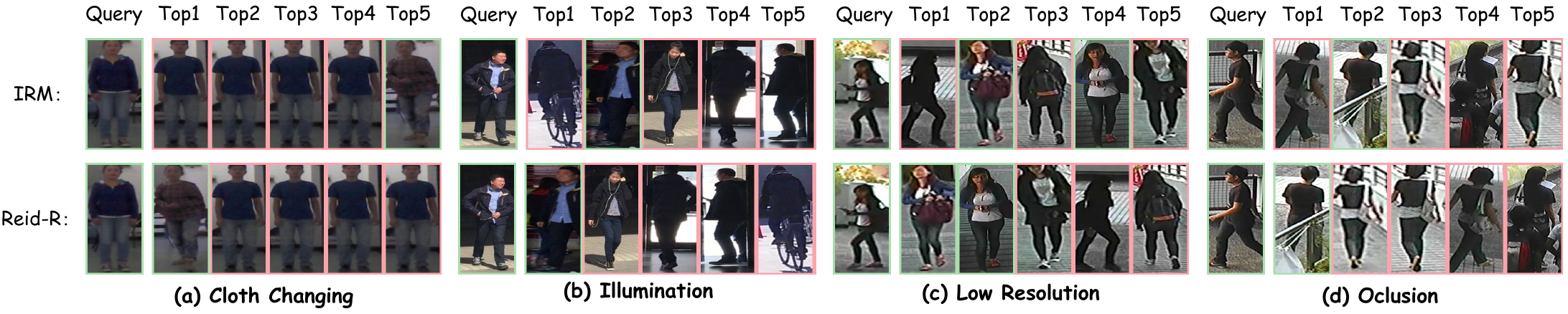}
    \caption{\textbf{Comparison of top-5 retrieval results from IRM~\cite{he2024instruct} and our ReID-R across four challenging scenarios.} For each scenario (a-d), green boxes denote correct matching, while red boxes denote false matching.}
    \label{fig:qualitative_ranking}
\end{figure*}

\begin{figure*}[t]
    \centering
    \includegraphics[width=\linewidth]{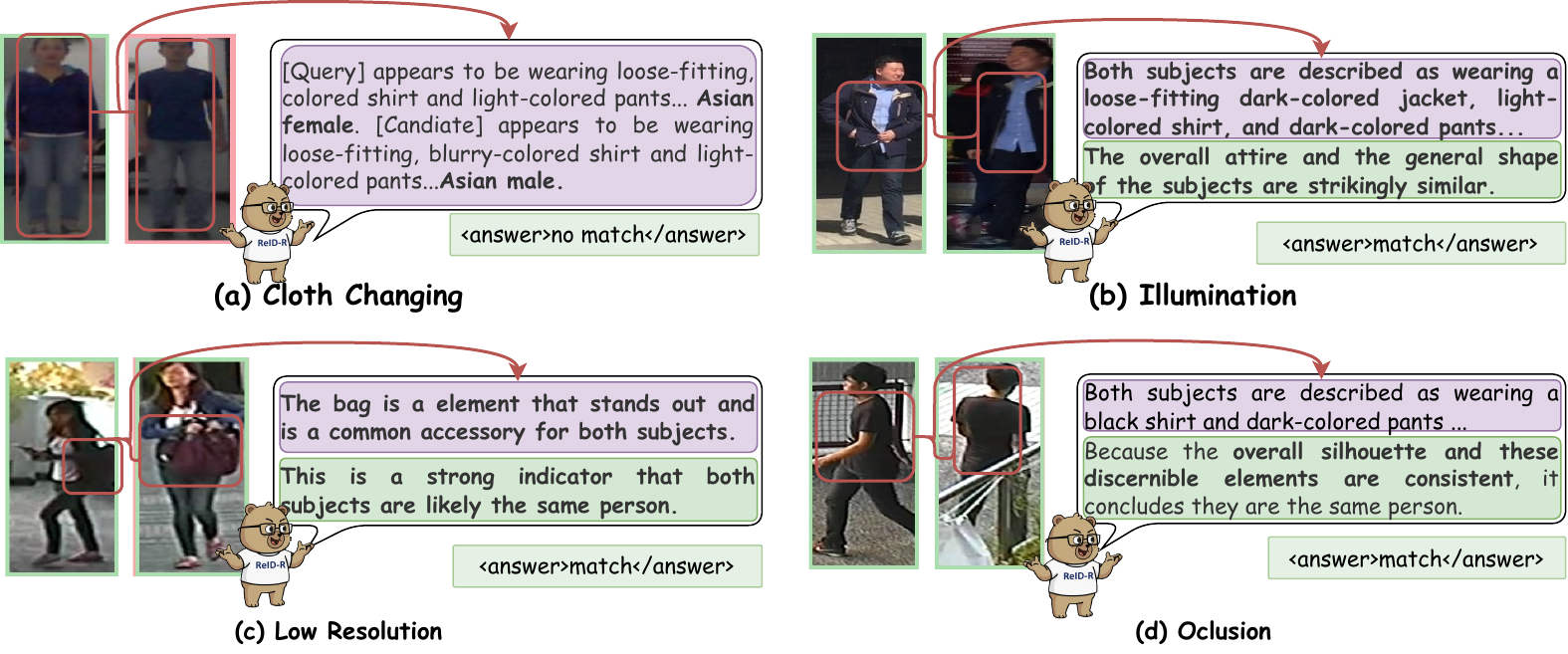}
    \caption{Qualitative analysis of the reasoning process generated by ReID-R.}
    \label{fig:qualitative_reasoning}
    \vspace{-1mm}
\end{figure*}
\subsection{Experimental Setup}
\paragraph{\textbf{Datasets and Metrics.}} We evaluate the performance of our model, ReID-R, on the test sets of five widely-used ReID benchmarks. These datasets span both traditional ReID tasks, including Market1501~\cite{zheng2015person}, MSMT17~\cite{wei2018person}, and CUHK03~\cite{li2014deepreid}, as well as challenging clothes-changing ReID tasks, namely PRCC~\cite{yang2019person} and VC-Clothes~\cite{wan2020person}. Following standard evaluation protocols, we report the Rank-1 accuracy and Mean Average Precision (mAP) as our primary metrics. In practice, our framework relies on a pairwise VQA formulation, which presents a challenge at inference time: processing the entire gallery for a single query is infeasible due to the MLLM's context limitations. To address this challenge, we implement a multi-step retrieval pipeline. We first adopt a standard ViT-based retrieval model to filter the gallery. Only the top-5 most similar gallery candidates are retained for each query. This step significantly reduces the number of candidates required for fine-grained processing. Next, we reformulate the retrieval task for our ReID-R. Each of the five selected candidates is individually paired with the query, reformulating the task as a sequential, pairwise VQA problem. Finally, our ReID-R makes a binary judgment for each of the five pairs. The final similarity list is produced by sorting the candidates judged to be matches ahead of others.

\paragraph{\textbf{Implementation Details.}} ReID-R, is initialized from the weights of Qwen2.5-VL-7B-Instruct~\cite{bai2025qwen2}, which is trained using 2 NVIDIA L40 GPUs, each with 48GB of VRAM. To ensure computational efficiency during training, we employ Low Rank Adaptation~\cite{hu2022lora} (Rank$=8$, $\alpha=16$, dropout=$0.2$). As is common practice, all images are resized to 256$\times$128 during both training and inference. For the GRPO optimization, the $G$ in Eq.~\ref{eq:grpoloss} is set to 8, with learning rate 5e-5. Further comprehensive details are provided in the Appendix.

\subsection{Comparison with State-of-the-Arts}
\paragraph{\textbf{Standard Person Re-ID}}

As shown in Table~\ref{tab:tra-ReID}, ReID-R demonstrates highly competitive performance across three widely used traditional Person Re-ID benchmarks. Significantly, our model establishes a new state of the art on the CUHK03 dataset, achieving 85.6\% mAP and 87.7\% Rank-1 accuracy. This result surpasses the previous best model IRM by a substantial margin of 2.3\% mAP and 1.2\% Rank-1. On the large-scale Market1501 and MSMT17 datasets, ReID-R consistently delivers results comparable to SOTA methods such as PASS and IRM. For instance, on Market1501, ReID-R achieves 92.5\% mAP and 96.5\% Rank-1, outperforming the IRM baseline. Furthermore, while MSMT17 presents a more challenging environment, our model maintains a robust Rank-1 accuracy of 86.8\%, which is superior to the IRM.

The strong performance largely stems from the robust discriminative capabilities of ReID-R in complex real world scenarios, as shown in Fig.~\ref{fig:qualitative_ranking}(b). Under severe illumination changes, the black box baseline is easily misled by spurious visual overlaps, such as the dark silhouette of a different person riding a bicycle, and retrieves a False Positive at the Top 1 rank. In contrast, ReID-R correctly identifies the true match at the first rank, indicated by the green box, by grounding its reasoning in stable identity evidence. This ability to effectively suppress environmental noise and focus on invariant identity features demonstrates superior robustness, confirming that ReID-R not only matches but in certain critical metrics exceeds the capabilities of current methods.

\begin{figure*}[t]
    \centering
    \includegraphics[width=\linewidth]{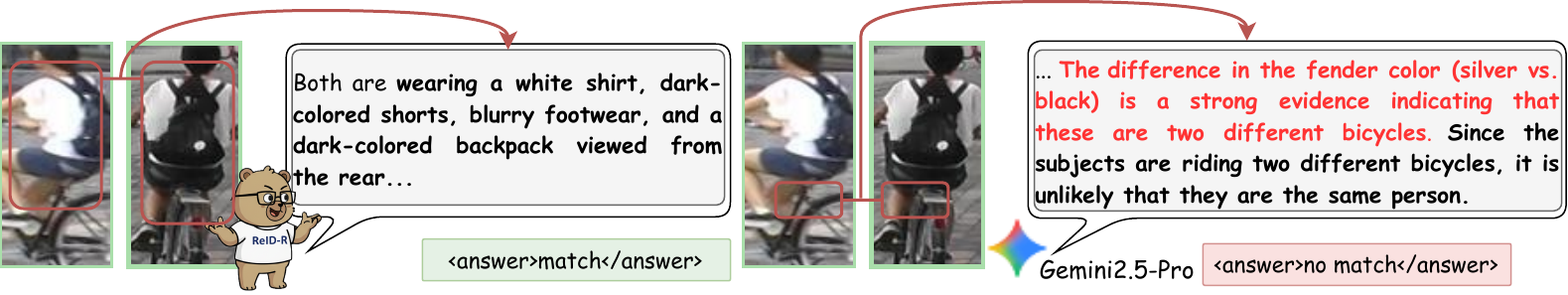}
    \caption{Visualizing object-identity confusions. While Gemini~2.5~Pro incorrectly treats the bicycle as identity evidence. In contrast, ReID-R correctly focuses on identity-related cues like clothes and backpack to reach the correct conclusion.}
    \label{fig:gemini_vs_reidr}
\end{figure*}

\begin{table}[!t]
  \centering
  \setlength{\tabcolsep}{1pt}
  \renewcommand\arraystretch{0.9}
  \caption{Ablation study on Market1501 and CUHK03 datasets, where ``NTS'' means our non-trivial sampling.}
    \begin{tabular}{lccccccc}
    \toprule
    \multirow{2}{*}{\textbf{Base Model}} & \multicolumn{3}{c}{\textbf{Components}} & \multicolumn{2}{c}{\textbf{Market1501}} & \multicolumn{2}{c}{\textbf{CUHK03}} \\
    \cmidrule(lr){2-4} \cmidrule(lr){5-6} \cmidrule(lr){7-8}
    & NTS & Stage 1 & Stage 2 & mAP & Rank-1 & mAP & Rank-1 \\
    \midrule
    Qwen2.5VL-7B~\cite{bai2025qwen2}  &      &      &      & 92.4 & 96.3& 84.0 & 85.6 \\
    Gemini 2.5 Pro~\cite{comanici2025gemini} & & & &91.4 &94.6 &78.0 &77.6 \\
    \midrule
    Qwen2.5VL-7B~\cite{bai2025qwen2} &      &      & $\checkmark$ &   \multicolumn{4}{c}{Mode Collapse (always 0)}\\
    Qwen2.5VL-7B~\cite{bai2025qwen2} & $\checkmark$ &      & $\checkmark$ &  92.2 &95.9 &83.9 &83.4\\
    Qwen2.5VL-7B~\cite{bai2025qwen2} & $\checkmark$ & $\checkmark$ &      & 92.4 & 96.2 & 83.3 & 82.7 \\
    Qwen2.5VL-7B~\cite{bai2025qwen2} & $\checkmark$ & $\checkmark$ & $\checkmark$ & \textbf{92.5} & \textbf{96.5} & \textbf{85.6} & \textbf{87.7} \\
    \bottomrule
    \end{tabular}
  \label{tab:ablation}
\end{table}

\begin{figure*}[t]
    \centering
    \includegraphics[width=\linewidth]{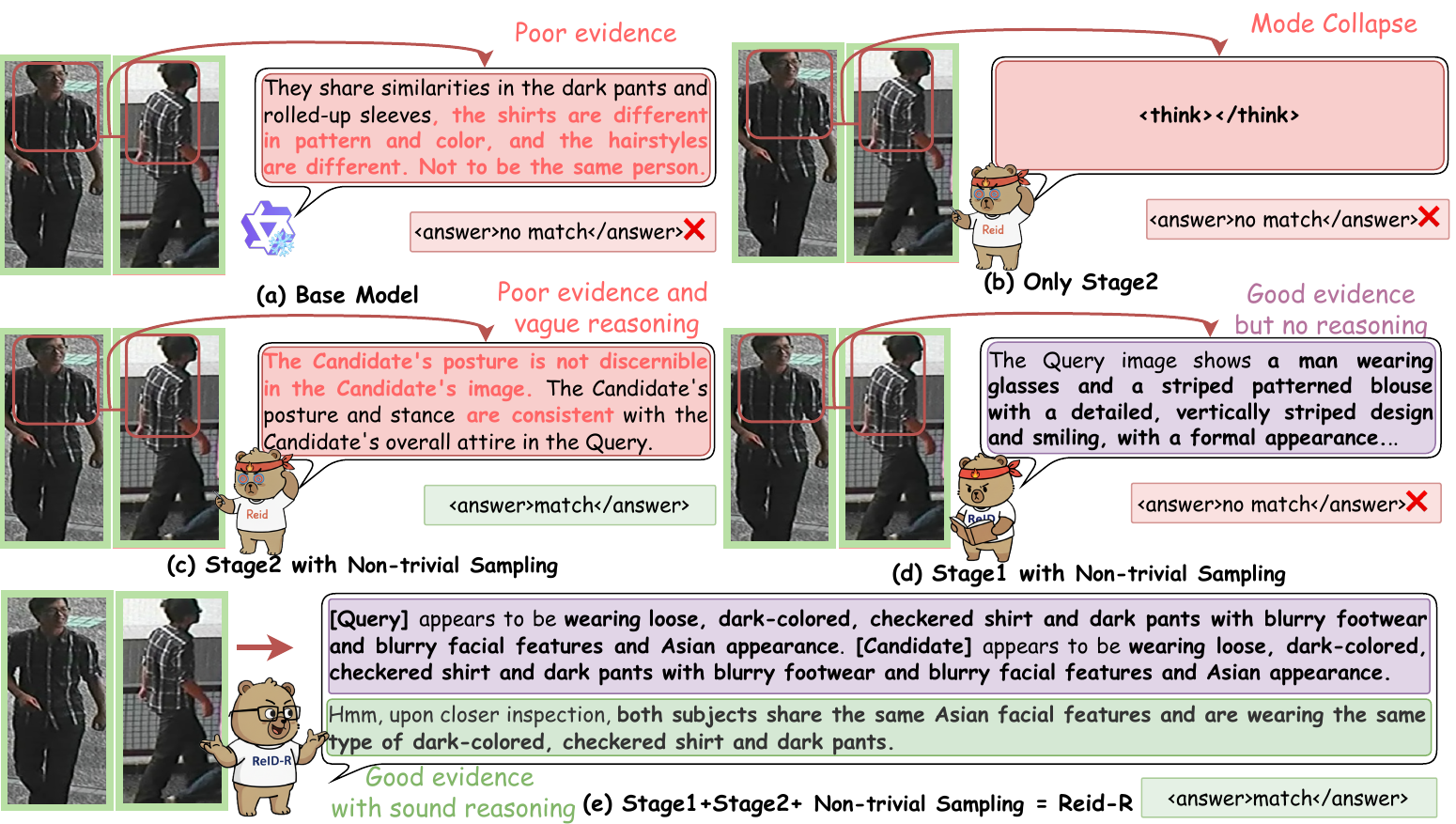}
    \caption{Qualitative insights from our ablation study. (a) The \textbf{Base Model} extracts poor evidence leading to an incorrect judgment. (b) \textbf{Only Stage 2} causes complete mode collapse. (c) For \textbf{Stage 2 with Nontrivial Sampling}, the model concludes they are the same person based on matching clothing, but its reasoning is logically flawed and contradictory. (d) \textbf{Stage 1 with Nontrivial Sampling} accurately describes features but lacks explicit logical deduction. (e) Our \textbf{Full ReID-R Model} seamlessly integrates accurate evidence discovery with sound logical reasoning.}
    \label{fig:ablation_viz}
\end{figure*}

\paragraph{\textbf{Cloth-changing Re-ID (CC Re-ID). }}
The results in Tab.~\ref{tab:CC-ReID} further confirm the effectiveness of our model in the highly challenging cloth changing scenario. On the PRCC dataset, ReID-R establishes the highest mAP of 46.2\%, effectively surpassing the strong IRM baseline. This competitive performance is also evident on the VC-Clothes dataset, where our model achieves the leading Rank-1 accuracy of 90.2\%. These quantitative gains are visually corroborated in Fig.~\ref{fig:qualitative_ranking}(a), where a clear pattern emerges. While the IRM model heavily overfits to surface-level features and is frequently confused by apparel variations, ReID-R successfully avoids this trap. Our model does not rely on dominant but changeable visual cues such as clothing color. Instead, by reasoning over identity-related features, ReID-R ensures accurate retrieval even when superficial appearances change, demonstrating a robust capability to extract invariant identity representations across diverse environments.

\subsection{Qualitative Analysis and Interpretability}

Figure~\ref{fig:qualitative_reasoning} exposes the explicit reasoning underpinning these improved decisions, indicating that our model has genuinely learned the core matching cues for person Re-ID. The top row highlights its remarkable ability to resist misleading visual overlaps and severe environmental noise. In (a) Cloth Changing, the model correctly predicts a negative match despite a highly deceptive visual setup. While it observes that both subjects are wearing seemingly identical loose-fitting shirts and light colored pants, it refuses to rely on these superficial clothing matches. Instead, it successfully isolates the fundamental biological difference, explicitly noting that the query is an Asian female while the candidate is an Asian male. Similarly, in (b) Illumination, heavy shadows completely obscure the facial details of the subject. The model overcomes this by assembling a comprehensive structural profile. It explicitly identifies the combination of a loose fitting dark colored jacket and a light colored shirt, and it further grounds its correct match prediction by observing that the general shape of the subjects are strikingly similar. Moreover, the bottom row further demonstrates robust reasoning under severe visual degradation. In (c) Low Resolution, where fine details are lost, the model correctly predicts a match by capturing a highly discriminative accessory, reasoning that the prominent bag is a strong indicator that both subjects are likely the same person. Finally, in (d) Occlusion, despite heavy viewpoint obstructions, the model successfully links the consistent attire and relies on the overall silhouette to confidently deduce a match.

Overall, these qualitative results suggest that ReID-R is not an opaque black box. Its predictions are supported by interpretable, step-by-step comparisons. This behavior follows naturally from our two-stage framework. Stage 1 trains the model to discover discriminative cues, such as a specific accessory or consistent body silhouette, while suppressing spurious environmental factors. Stage 2 then supervises the model to compose these cues into a coherent rationale that culminates in the final matching decision.

\subsection{Ablation Study}
To validate the necessity of each proposed component and the underlying mechanisms of our two stage framework, we conduct extensive ablation studies on the Market1501 and CUHK03 datasets. The quantitative results are presented in Table~\ref{tab:ablation}. By correlating these empirical results with our theoretical formulations, we demonstrate exactly how our framework transforms a general vision language model into a robust Re-ID expert.

\paragraph{\textbf{Comparison with other MLLMs.}} We evaluate the zero-shot performance of the base Qwen2.5VL-7B model against a strong closed-source MLLM, Gemini 2.5 Pro. Despite its massive scale, Gemini 2.5 Pro significantly underperforms on the challenging CUHK03 dataset, achieving only 78.0\% mAP compared to 84.0\% for the base model and 85.6\% for our final ReID-R. As corroborated by Figure~\ref{fig:gemini_vs_reidr}, general-purpose MLLMs lack domain-adaptive identity reasoning. They erroneously treat transient contextual cues, such as the color of a bicycle, as definitive identity evidence. This misplaced reliance on changeable attributes inevitably leads to spurious rationales and incorrect matches. These results firmly confirm that simply scaling general capabilities is insufficient, and effective person Re-ID strictly requires targeted mechanisms to extract stable identity evidence.

\paragraph{\textbf{Necessity of NTS for Reward Stability.}} We further ablate the key training mechanisms in Table~\ref{tab:ablation}. Applying Stage 2 directly without Stage 1 and NTS causes complete mode collapse (Figure~\ref{fig:ablation_viz}b), where the model fails to generate any reasoning. This stems from a severe data imbalance. In randomly constructed datasets, an overwhelming 99.83\% of pairs are negative. Confronted with this 0.17\% positive disparity, the model exploits the reward system by simply outputting ``0'' for all instances to bypass genuine reasoning. Introducing NTS balances this distribution and successfully avoids collapse, achieving 83.9\% mAP on CUHK03. However, as Figure~\ref{fig:ablation_viz}(c) illustrates, without the Stage 1 warm up, the reasoning remains logically flawed. The model confusingly states that the Candidate's clothing matches the Query, while simultaneously claiming the posture is both not discernible and consistent.

\paragraph{\textbf{Synergy of Two Stage Optimization.}} Using only the Stage 1 Discriminative Captioning yields a stable 83.3\% mAP on CUHK03. As shown in Figure~\ref{fig:ablation_viz}(d), this stage successfully primes the model to accurately extract identity evidence, although it lacks explicit logical deduction. Furthermore, Stage 1 relies on the strong prior that a positive sample exists within the gallery. To achieve independent pairwise verification for real world applications, Stage 2 is strictly required. When both stages are combined with NTS, performance leaps to 85.6\% mAP and 87.7\% Rank-1 on CUHK03. As corroborated by Figure~\ref{fig:ablation_viz}(e), this perfectly highlights the complementarity of our framework. Stage 1 autonomously grounds high quality evidence, while Stage 2 explicitly optimizes the deductive process to convert that evidence into accurate decisions.

\section{Conclusion}
In this work, we propose ReID-R, a novel reasoning-driven paradigm that successfully reformulates person re-identification as a step-by-step reasoning task. Specifically, we design a two-stage reinforcement learning framework that contains: (1) Discriminative Reasoning Warm-up trains a policy model to autonomously acquire identity-aware feature understanding. (2) Efficient Reinforcement Learning designed a non-trivial sampling to guarantee high-quality reward signals in training, enabling accurate and scene-generalizable reasoning. Extensive experiments demonstrate that ReID-R achieves competitive performance, using only 14.3K non-trivial data points, representing just 20.9\% of the data scale. Furthermore, as a direct benefit of its inherent design, ReID-R generates high-quality, verifiable interpretations for its matching results.

\newpage


\bibliographystyle{ACM-Reference-Format}
\bibliography{arxiv}

\end{document}